\documentclass[letterpaper]{article}
\usepackage{aaai18}
\usepackage{times}
\usepackage{helvet}
\usepackage{courier}
\usepackage{url}
\usepackage{graphicx}
\graphicspath{{images/}}
\usepackage{booktabs}       
\usepackage{amsfonts}       
\usepackage{nicefrac}       
\usepackage{microtype}      
\usepackage{subfigure}
\usepackage{amsmath}
\usepackage{bm}
\title{DarkRank: Accelerating Deep Metric Learning via Cross Sample Similarities Transfer}
\author{
	Yuntao Chen$^{1,5}$ \quad Naiyan Wang$^{2}$ \quad Zhaoxiang Zhang$^{1,3,4,5,}$\thanks{corresponding author}\\
	$^{1}$Research Center for Brain-inspired Intelligence, CASIA \quad	$^{2}$TuSimple\\
	$^{3}$National Laboratory of Pattern Recognition, CASIA\\
	$^{4}$Center for Excellence in Brain Science and Intelligence Technology, CAS\\
	$^{5}$University of Chinese Academy of Sciences\\
	\{chenyuntao2016, zhaoxiang.zhang\}@ia.ac.cn \quad winsty@gmail.com
}
\def\C{{\bf C}}

\def\X{{\bf X}}

\def\q{{\bf q}}

\def\x{{\bf x}}

\def\0{{\bf 0}}
\def\1{{\bf 1}}

\def\MP{{\mathcal P}}

\def\BR{{\mathbb R}}

\def\etal{{\em et al.\/}\,}

\def\exp{\mathrm{exp}}

\newcommand{\norm}[1] {\|#1\|_2}

\begin{document}
	\maketitle
	\begin{abstract}
We have witnessed rapid evolution of deep neural network architecture design in the past years. These latest progresses greatly facilitate the developments in various areas such as computer vision and natural language processing. However, along with the extraordinary performance, these state-of-the-art models also bring in expensive computational cost. Directly deploying these models into applications with real-time requirement is still infeasible. 
Recently, Hinton \etal\cite{Hinton2015DistillingTK} have shown that the dark knowledge within a powerful teacher model can significantly help the training of a smaller and faster student network. These knowledge are vastly beneficial to improve the generalization ability of the student model. Inspired by their work, we introduce a new type of knowledge -- cross sample similarities for model compression and acceleration. This knowledge can be naturally derived from deep metric learning model. To transfer them, we bring the ``learning to rank'' technique into deep metric learning formulation. 
We test our proposed DarkRank method on various metric learning tasks including pedestrian re-identification, image retrieval and image clustering. The results are quite encouraging. Our method can improve over the baseline method by a large margin. Moreover, it is fully compatible with other existing methods. When combined, the performance can be further boosted.
	\end{abstract}

\section{Introduction}
Metric learning is the basis for many computer vision tasks, including face verification\cite{Schroff_2015,Taigman2014DeepFaceCT} and pedestrian re-identification\cite{Wang2016JointLO,Chen2015RelevanceML}. In recent years, end-to-end deep metric learning method which learns feature representation by the guide of metric based losses has achieved great success\cite{Qian2015FinegrainedVC,Song2016DeepML,Schroff_2015}. A key factor for the success of these deep metric learning methods is the powerful network architectures\cite{Xie2016,He2016DeepRL,Szegedy2015GoingDW}.
Nevertheless, along with more powerful features, these deeper and wider networks also bring in heavier computation burden. In many real-world applications like autonomous driving, the system is latency critical with limited hardware resources. To ensure safety, it requires (more than) real-time responses. This constraint prevents us from benefiting from the latest developments in network design.

To mitigate this problem, many model acceleration methods have been proposed. They can be roughly categorized into three types: network pruning\cite{NIPS1989_250,han2015learning}, model quantization\cite{NIPS2016_6573,Rastegari2016} and knowledge transfer\cite{zagoruyko2016paying,Romero2014FitNetsHF,Hinton2015DistillingTK}. Network pruning iteratively removes the neurons or weights that are less important to the final prediction; model quantization decreases the representation precision of  weights and activations in a network, and thus increases  computation throughput; knowledge transfer directly trains a smaller student network guided by a larger and more powerful teacher. 
Among these methods, knowledge transfer based methods are the most practical. Compared with other methods that mostly need tailor made hardwares or implementations, they can archive considerable acceleration without bells and whistles.

\begin{figure*}[t]
	\centering
	\makebox[\textwidth]{\includegraphics[width=1\textwidth]{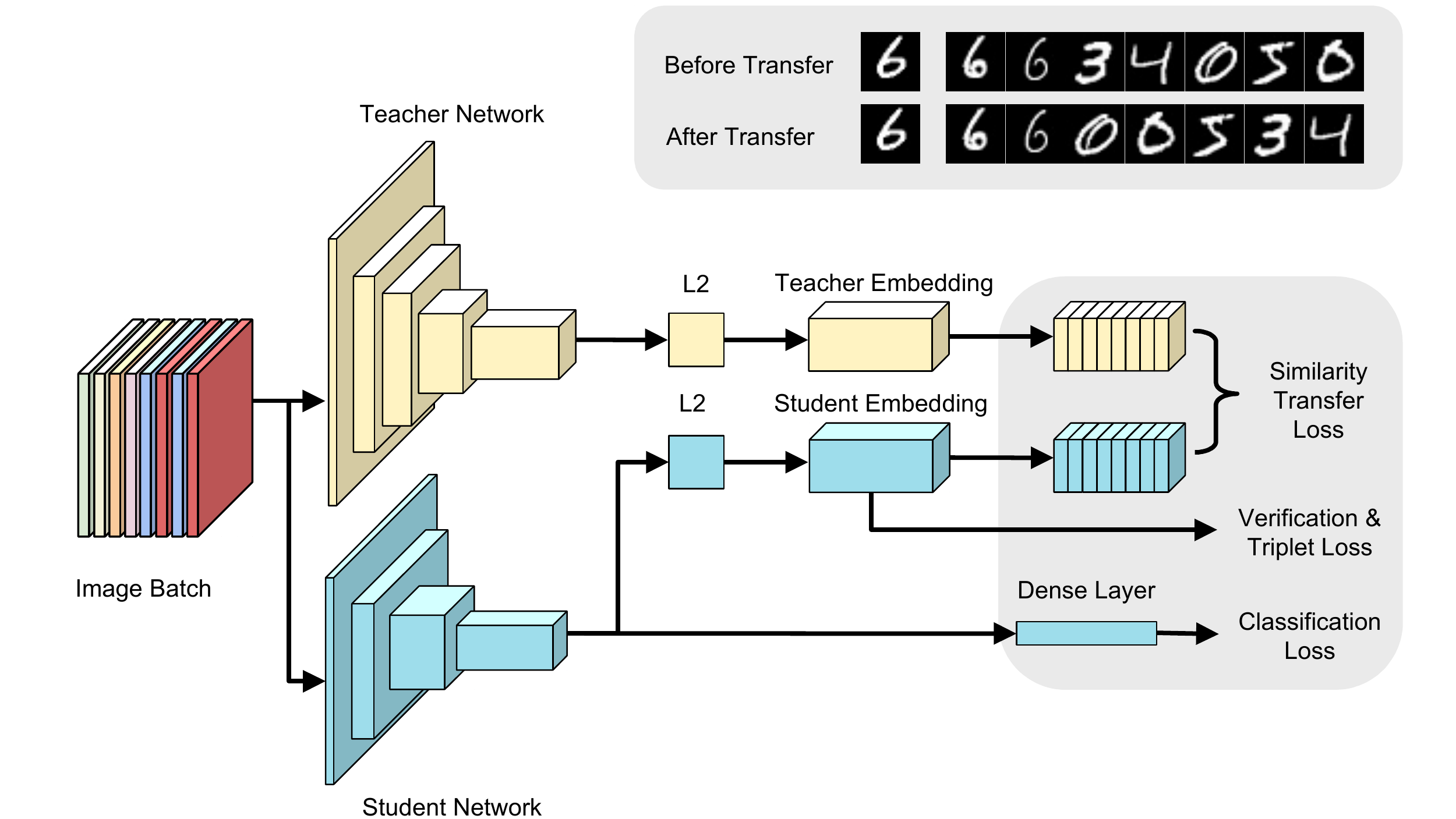}}
	\caption {The network architecture of our DarkRank method. The student network is trained with standard classification loss, contrastive loss and triplet loss as well as the similarity transfer loss proposed by us.}
	\label{fig:distill}
\end{figure*}

Knowledge Distill (KD)\cite{Hinton2015DistillingTK} and its variants\cite{zagoruyko2016paying,Romero2014FitNetsHF} are the dominant approaches among knowledge transfer based methods. Though they utilize different forms of knowledges, these knowledges are still limited within a single sample. Namely, these methods provide more precise supervision for each sample from teacher networks at either classifier or intermediate feature level. However, all these methods miss another valuable treasure -- the relationships  (similarities or distances) across different samples. This kind of knowledge also encodes the structure of the embedded space of teacher networks. Moreover, it naturally fits the objective of metric learning since it usually utilizes similar instance level supervision. We elaborate our motivation in the sequel, and depict our method in Fig.~\ref{fig:distill}. The upper right corner shows that the student better captures the similarity of images after transferring. The digit 0 which are more similar to 6 than 3, 4, 5 are now ranked higher.


To summarize, the contributions of this paper are three folds:

\begin{itemize}
	\item We introduce a new type of knowledge -- cross sample similarities for knowledge transfer in deep metric learning.
	\item We formalize it as a rank matching problem between teacher and student networks, and modify classical listwise learning to rank methods\cite{Cao2007LearningTR,xia2008listwise} to solve it.
	\item We test our proposed method on various metric learning tasks. Our method can significantly improve the performance of student networks. And it can be applied jointly with existing methods for a better transferring performance.
\end{itemize}

\section{Related works}
In this section, we review several previous works that are closely related to our proposed method.

\subsection{Deep Metric Learning}
Different from most traditional metric learning methods that focus on learning a Mahalanobis distance in Euclidean space\cite{NIPS2002_2164,Kwok:2003:LIK} or high dimensional kernel space\cite{weinberger2006distance}, deep metric learning usually transforms the raw features via DNNs, and then compare the samples in Euclidean space directly.

Despite the rapid evolution of network architectures, the loss functions for metric learning are still a popular research topic. The key point of metric learning is to separate inter-class embeddings and reduce the intra-class variance. Classification loss and its variants\cite{Liu2016LargeMarginSL,Wen2016} can learn robust features that help to separate samples in different classes. However, for out-of-sample identities, the performance cannot be guaranteed since no explicit metric is induced by this approach. 
Another drawback of classification loss is that it projects all samples with the same label to the same direction in the embedding space, and thus ignores the intra-class variance. Verification loss\cite{Bromley1993SignatureVU} is a popular alternative because it directly encodes both the similarity ans dissimilarity supervisions. 
The weakness of verification loss is that it tries to enforce a hard margin between the anchor and negative samples. This restriction is too strict since images of different categories may look very similar to each other. Imposing a hard margin on those samples only hurts the learnt representation. Triplet loss and its variants\cite{Cheng2016PersonRB,Liu2016MultiScaleTC} overcome this disadvantage by imposing an order on the embedded triplets instead. Triplet loss is the exact reflection of desired retrieval results: the positive samples are closer to anchor than the negative ones. But its good performance requires a careful design of the sampling and the training procedure\cite{Schroff_2015,Hermans2017InDO}. Other related work includes center loss~\cite{Wen2016} which maintains a shifting template for each class to reduce the intra-class variance by simultaneously drawing the template and the sample towards each other.  Besides loss function design, Bai \etal \cite{Bai_2017_CVPR} introduce smoothness of metric space with respect to data manifold as a prior. 

\subsection{Knowledge Transfer for Model Acceleration and Compression}
In \cite{bucila2006model}, Bucila \etal first proposed to approximate an ensemble of classifiers with a single neural network. Recently, Hinton \etal revived this idea under the name knowledge distill\cite{Hinton2015DistillingTK}. The insight comes from that the softened probabilities output by classifiers encode more accurate embedding of each sample in the label space than one-hot labels. Consequently, in addition to the original training targets, they proposed to use soft targets from teacher networks to guide the training of student networks. Through this process, KD transfers more precise supervision signal to student networks, and therefore improves their generalization ability. Subsequent works FitNets\cite{Romero2014FitNetsHF}, Attention Transfer\cite{zagoruyko2016paying} and Neuron Selectivity Transfer\cite{huang2017nst} tried to exploit other knowledges in intermediate feature maps of CNNs to improve the performance. Instead of using forward input-output pairs, Czarnecki \etal tried to utilize the gradients with respect to input of teacher network for knowledge transfer with Sobolev training\cite{czarnecki2017sobolev}. In this paper, we exploit a unique type of knowledge inside deep metric learning model -- cross sample similarities to train a better student network.

\subsection{Learning to Rank}
Learning to rank refers to the problem that given a query, rank a list of samples according to their similarities. Most learning to rank methods can be divided into three types: pointwise, pairwise and listwise, according to the way of assembling samples. Pointwise approaches~\cite{cossock2006subset,shashua2003ranking} directly optimize the relevance label or similarity score between the query and each candidate; while pairwise approaches compare the relative relevance or similarity of two candidates. Representative works of pairwise ranking include Ranking SVM~\cite{herbrich2000large} and Lambda Rank~\cite{Burges2006LearningTR}. Listwise methods either directly optimize the ranking evaluation metric or maximize the likelihood of the ground-truth rank. SVM MAP~\cite{yue2007support}, ListNet~\cite{Cao2007LearningTR} and ListMLE~\cite{xia2008listwise} fall in this category. In this paper, we introduce listwise ranking loss into deep metric learning, and utilize it to transfer the soft similarities between candidates and the query into student models.

\section{Background}
In this section, we review ListNet and ListMLE which are classical listwise learning to rank methods introduced by Cao \etal\cite{Cao2007LearningTR} and Xia \etal\cite{xia2008listwise} for document retrieval task. These methods are closely related to our proposed method that will be elaborated in the sequel.

The core idea of these methods is to associate a probability with every rank permutation based on the relevance or similarity score between candidate $\x$ and query $\q$. 

We use $\pi$ to denote a permutation of the list indexes. For example, a list of four samples can have a permutation of $\pi = \left\{\pi(1), \pi(2), \pi(3), \pi(4)  \right\} = \left\{ 4, 3, 1, 2 \right\}$, which means the forth sample in the list is ranked first, the third sample second, and so on.
Formally, We denote the candidate samples as $\X \in \BR^{p \times n}$ with each column $i$ being a sample $\x_i \in \BR^{p}$. Then the probability of a specific permutation $\pi$ is given as:
\begin{equation}
\label{eqn:permprob}
P(\pi|\X)= \prod_{i=1}^{n}\dfrac{\exp[S(\x_{\pi(i)})]}{\sum_{k=i}^{n}\exp[S(\x_{\pi(k)})]}
\end{equation}
where $S(\x)$ is a score function based on the distance between $\x$ and $\q$.
After the probability of a single permutation is constructed, the objective function of ListNet can be defined as:
\begin{equation}
L_\text{ListNet}(\x) = -\sum_{\pi\in\mathcal{P}}P(\pi|\mathbf{s})\log P(\pi|\mathbf{x})
\end{equation}
where $\mathcal{P}$ denotes all permutations of a list of length $n$, and $\mathbf{s}$ denotes the ground-truth.

Another closely related method is ListMLE\cite{xia2008listwise}. Unlike ListNet, as its name states, ListMLE aims at maximizing the likelihood of a ground truth ranking $\pi_y$. The formal definition is as follow:
\begin{equation}
L_\text{ListMLE}(\x) = -\log P(\pi_y|\x)
\end{equation} 

\section{Our Method}
In this section, we first introduce the motivation of our DarkRank by an intuitive example, then followed by the formulation and two variants of our proposed method. 
\subsection{Motivation}
We depict our framework in Fig.~\ref{fig:distill} along with an intuitive illustration to explain the motivation of our work. In the example, the query is a digit 6, and there are two relevant digits and six irrelevant digits. Through training with such supervision, the original student network can successfully rank the relevant digits in front of the irrelevant ones. However, for the query 6, there are two 0s which are more similar than other digits. Simply using hard labels (similar or dissimilar) totally ignores such dark knowledge. However, such knowledge is crucial for the generalization ability of student models. A powerful teacher model may reflect these similarities its the embedded space. Consequently, we propose to transfer these cross sample similarities to improve the performance of student networks.

\subsection{Formulation}
We denote the embedded features of each mini-batch after an embedding function $f(\cdot)$ as $\X$. Here the choice of $f(\cdot)$ depends on the problem at hand, such as CNN for image data or DNN for text data. We further use $\X^s$ to denote the embedded features from student networks, and similarly $\X^t$ for those from teacher networks.
We use one sample in the mini-batch as the anchor query $\q = \x_1$, and the rest samples in the mini-batch as candidates $\C = \{\x_2, \x_3, \cdots, \x_n\}$. 
We then construct a similarity score function $S(\x)$ based on the Euclidean distance between two embeddings. The $\alpha$ and $\beta$ are two parameters in the score function to control the scale and ``contrast'' of different embeddings:
\begin{equation}
S(\x) = -\alpha \|\q-\x\|_2^\beta.
\end{equation}

After that, we propose two methods for the transfer: soft transfer and hard transfer. For soft transfer method, we construct two probability distributions $P(\pi\in \MP \mid \X^s)$ and $P(\pi\in \MP \mid \X^t)$ over all possible permutations (or ranks) $\MP$ of the mini-batch based on Eqn.~\ref{eqn:permprob}. Then, we match these two distributions with KL divergence. For hard transfer method, we simply maximize the likelihood of the ranking $\pi_y$ which has the highest probability by teacher model. Formally, we have

\begin{equation}
\begin{split}
L_\text{soft}(\X^s, \X^t) &= D_\text{KL}[P(\pi\in \MP \mid \X^t) ~\|~ P(\pi\in \MP \mid \X^s)] \\
&= \sum_{\pi \in \MP} P(\pi \mid \X^t) \log \dfrac{P(\pi \mid \X^t)}{P(\pi \mid \X^s)},\\
L_\text{hard}(\X^s, \X^t) &= -\log P(\pi_y \mid \X^s, \X^t).
\end{split}
\end{equation}

Soft transfer considers all possible rankings. It is helpful when there are several rankings with similar probability. However, there are $n!$ possible ranking in total. It is only feasible when $n$ is not too large. Whereas, hard transfer only considers the most possible ranking labeled by the teacher. As demonstrated in the experiments, hard transfer is a good approximation of soft transfer in the sense that it is much faster with long lists but has similar performance.

For the gradient calculation, we first use $S_i$ to denote $S(\x_{\pi(i)})$ for better readability, then the gradient is calculated as below:

\begin{equation}
\begin{split}
\dfrac{\partial P}{\partial S_i} &= \prod_{k=2}^{n}\dfrac{\exp(S_k)}{\sum_{m=k}^{n}\exp(S_m)} \\ &-\sum_{j=1}^{i}\left[\left(\prod_{k=2}^{n}\dfrac{\exp(S_k)}{\sum_{m=k}^{n}\exp(S_m)}\right)\dfrac{\exp(S_i)}{\sum_{m=j}^{n}\exp(S_m)}\right].
\end{split}
\end{equation}

For the gradient of $S_i$ with respect to $\x$, it is trivial to calculate. So we don't expand it here.

The overall loss function for the training of student networks consists both losses from ground-truth and loss from teacher knowledge. In specific, we combine large margin softmax loss~\cite{Liu2016LargeMarginSL}, verification loss~\cite{Bromley1993SignatureVU} and triplet loss~\cite{Schroff_2015} and the proposed DarkRank loss which can either be its soft or hard variant. 

\begin{figure*}[t]
	\centering
	\makebox[\textwidth]{\includegraphics[width=1.0\textwidth]{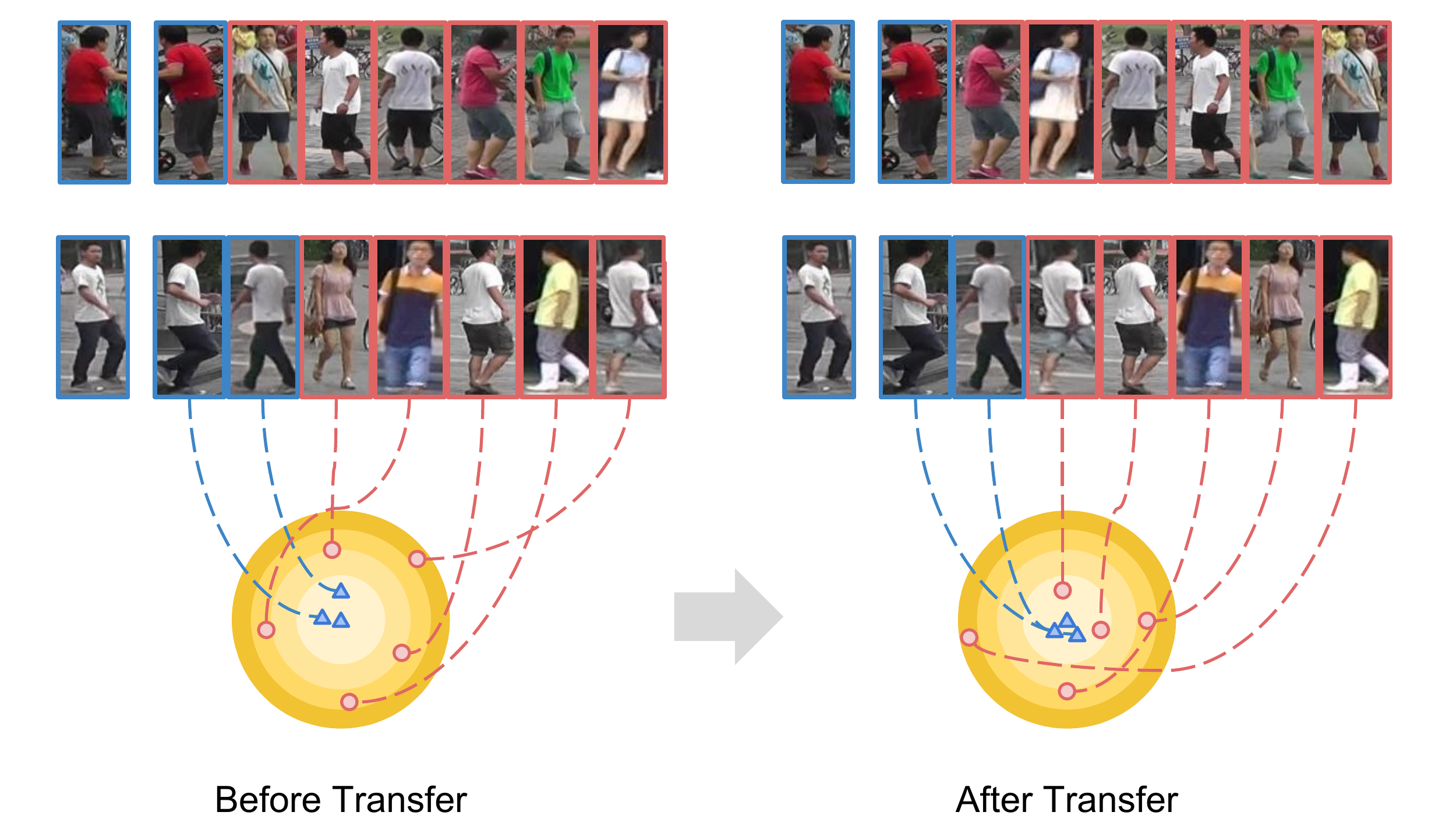}}
	\caption{Selected results visualization before and after our DarkRank transfer on Market1501. The border color of image denotes its relation to the query image. With the help of teacher's knowledge, the student model learns a better distance metric that can capture similarities in images.}
	\label{fig:showcase}
\end{figure*}

\section{Experiments}
In this section, we test the performance of our DarkRank method on several metric learning tasks including person re-identification, image retrieval and clustering,  and compare it with several baselines and closely related works. We also conduct ablation analysis on the influence of the hyper-parameters in our method. 
\subsection{Datasets}
We briefly introduce the datasets will be used in the following experiments.

\paragraph{CUHK03} CUHK03\cite{Li2014DeepReIDDF} is a large scale data for person re-identification. It contains 13164 images of 1360 identities. Each identity is captured by two cameras from different views. The author provides both detected and hand-cropped annotations. We conduct our experiments on the detected data since it is closer to the real world scenarios. Furthermore, we follow the training and evaluation protocol in\cite{Li2014DeepReIDDF}. We report Rank-1, 5 and 10 performance on the first standard split. 

\paragraph{Market1501} Market1501\cite{Zheng2015ScalablePR} contains 32668 images of 1501 identities. These images are collected from six different camera views. We follow the training and evaluation protocol in \cite{Zheng2015ScalablePR}, and report mean Average Precision (mAP) and Rank-1 accuracy in both single and multiple query settings. 

\paragraph{CUB-200-2011} The Caltech UCSD Birds-200-2011 (CUB-200-2011) dataset contains 11788 images of 200 bird species. Following the setting in \cite{Song2016DeepML}, we train our network on the first 100 species (5864 images) and then perform image retrieval and clustering on the rest 100 species (5924 images). Standard $F_1$, NMI and Recall@1 metrics are reported.

\subsection{Implementation Details}
We choose Inception-BN\cite{Szegedy2015GoingDW} as our teacher network and NIN-BN\cite{Lin2013NetworkIN} as our student network. Both networks are pre-trained on the ImageNet LSVRC image classification dataset\cite{ILSVRC15}. We first remove the fully connected layers specific to the pre-trained task, and then globally average pool the features. The output is then connected to a fully connected layer followed a L2 normalization layer to generate the final embeddings. The large margin softmax loss is directly connected to the fully connected layer. All other losses including the proposed transfer loss are built upon the L2 normalization layer. Figure~\ref{fig:distill} illustrates the architecture of our system. 

We set the margin in large margin softmax loss to 3, and set the margin to 0.9 in both triplet and verification loss. We set the loss weights of verification, triplet and large margin softmax loss to 5, 0.1, 1, respectively. We choose the stochastic gradient descent method with momentum for optimization. We set the learning rate to $0.01$ for the Inception-BN and $5\times 10^{-4}$ for the NIN-BN, and set the weight decay to $10^{-4}$. We train the model for 100 epochs, and shrink the learning rate by a factor of 0.1 at 50 and 75 epochs. The batch size is set to 8.

For person ReID tasks, we resize all input images to 256$\times$128 and randomly crop to 224$\times$112. We first construct all possible cross view positive image pairs, and randomly shuffle them at the start of each epoch. For image retrieval and clustering, we resize all input images to 256$\times$256 and randomly crop to 224$\times$224.  In addition, we flip the images in horizontal direction randomly during the training of both tasks. We implement our method in MXNet~\cite{chen2015mxnet}. We train our model from scratch when experimenting with CUB-200-2011 dataset, since the authors discourage the use of ImageNet pre-trianed model due to sample overlap. 

\subsection{Compared Methods}
We introduce the models and baselines compared in our experiments. Despite the soft and hard DarkRank methods proposed by us, we also test the following methods and the combination of them with our methods:

\paragraph{Knowledge Distill (KD)} Since the classification loss is included in our model, we test the knowledge distill with softened softmax target. According to \cite{Hinton2015DistillingTK}, we set the temperature $T$ to 4 and the loss weight to 4$^2$ for softmax knowledge distill method. Formally, KD can be defined as:

\begin{equation}
\begin{split}
&L_\text{KD}(\X^s,\X^t)=\\
&\sum_{i=1}^{n}D_\text{KL}\left[\text{softmax}\left(\dfrac{\x^t_i}{T}\right)\|\text{softmax}\left(\dfrac{\x^s_i}{T}\right)\right].
\end{split}
\end{equation}

\paragraph{Direct Match} Distances between the query and candidates are the most straightforward form of cross sample similarities knowledge. So we directly match the distances output by teacher and student models as a baseline. Formally, the matching loss is defined as:

\begin{equation}
L_\text{match}(\X^s,\X^t) = \sum_{i=2}^n\left(\norm{\x^s_i-\q^s}^2 - \norm{\x^t_i-\q^t}^2\right)^2.
\end{equation} 

\subsection{Person ReID Results}
We present the results of Market1501 and CUHK03 in Table.~\ref{table:Market1501} and Table.~\ref{table:CUHK03}, respectively. 

\begin{table}[th]
	\centering
	\begin{tabular}{ccccc}
		\toprule
		& \multicolumn{2}{c}{Single Query} & \multicolumn{2}{c}{Multiple Query} \\
		\cmidrule{2-3} \cmidrule{4-5}
		Method                    & mAP  & Rank 1 & mAP & Rank 1  \\
		\midrule
		Student          & 58.1 & 80.3 & 66.7 & 86.7 \\
		Direct Match      & 58.5 & 80.3 & 68.0 & 86.7 \\
		Hard DarkRank & {\bfseries63.5} & 83.0 & 71.2 & 87.4 \\
		Soft DarkRank  & 63.1 & {\bfseries83.6} & {\bfseries71.4} & {\bfseries88.8} \\
		\midrule
		KD & 66.7 & 86.0 & 75.1 & 90.4 \\
		KD + HardRank & {\bfseries68.5} & 86.6 & 76.3 & 90.3 \\
		KD + SoftRank & 68.2 & {\bfseries86.7} & {\bfseries76.4} & {\bfseries91.4} \\
		\midrule
		Teacher    & 74.3 & 89.8 & 81.2 & 93.7 \\
		\bottomrule
	\end{tabular}
	\caption{ mAP(\%) and Rank-1 accuracy(\%) on Market1501 of various methods. We use average pooling of features in multi-query test.}
	\label{table:Market1501}
\end{table}

\begin{table}[th]
	\centering
	\begin{tabular}{cccc}
		\toprule
		Method                    & Rank 1     & Rank 5    & Rank 10   \\
		\midrule
		Student          & 82.6 & 95.2 & 97.4     \\
		Direct Match      & 82.6 & 95.6 & 97.7     \\
		HardRank & 86.0 & {\bfseries97.5} & {\bfseries98.8}     \\
		SoftRank  & {\bfseries86.2 }& {\bfseries97.5} & 98.6     \\
		\midrule
		KD  & 87.8 & 97.5 & 98.7 \\
		KD + HardRank & 88.6 & {\bfseries98.2} & {\bfseries99.0}  \\
		KD + SoftRank & {\bfseries88.7} & 98.0 & {\bfseries99.0} \\
		\midrule
		Teacher    & 89.7 & 98.4 & 99.2     \\
		\bottomrule
	\end{tabular}
	\caption{ Rank-1,5,10 accuracy(\%) of various methods on CUHK03.}
	\label{table:CUHK03}
\end{table}

From Table.~\ref{table:Market1501}, we can see that directly matching the distances between teacher and student model only has marginal improvement over the original student model. We owe the reason to that the student model struggles to match the exact distances as teacher's due to its limited capacity. As for our method, both soft and hard variants make significant improvements over the original model. They could get similar satisfactory results. As discussed in the formulation, the hard variant has great computational advantage over the soft one in training, thus it is more preferable for the practitioners. 
Moreover, in synergy with KD, the performance of the student model can be further improved. This complementary results demonstrate that our method indeed transfers the inter-instance knowledge in the teacher network which is ignored by KD. 

On CUHK03 dataset, we can observe similar trends as on Market1501, except that the model performance on CUHK03 is much higher, which makes the performance improvement less significant.

\subsection{Ablation Analysis}
\begin{figure*}[t]
	\subfigure[contrast in score function $\beta$]{
		\includegraphics[width=0.31\textwidth]{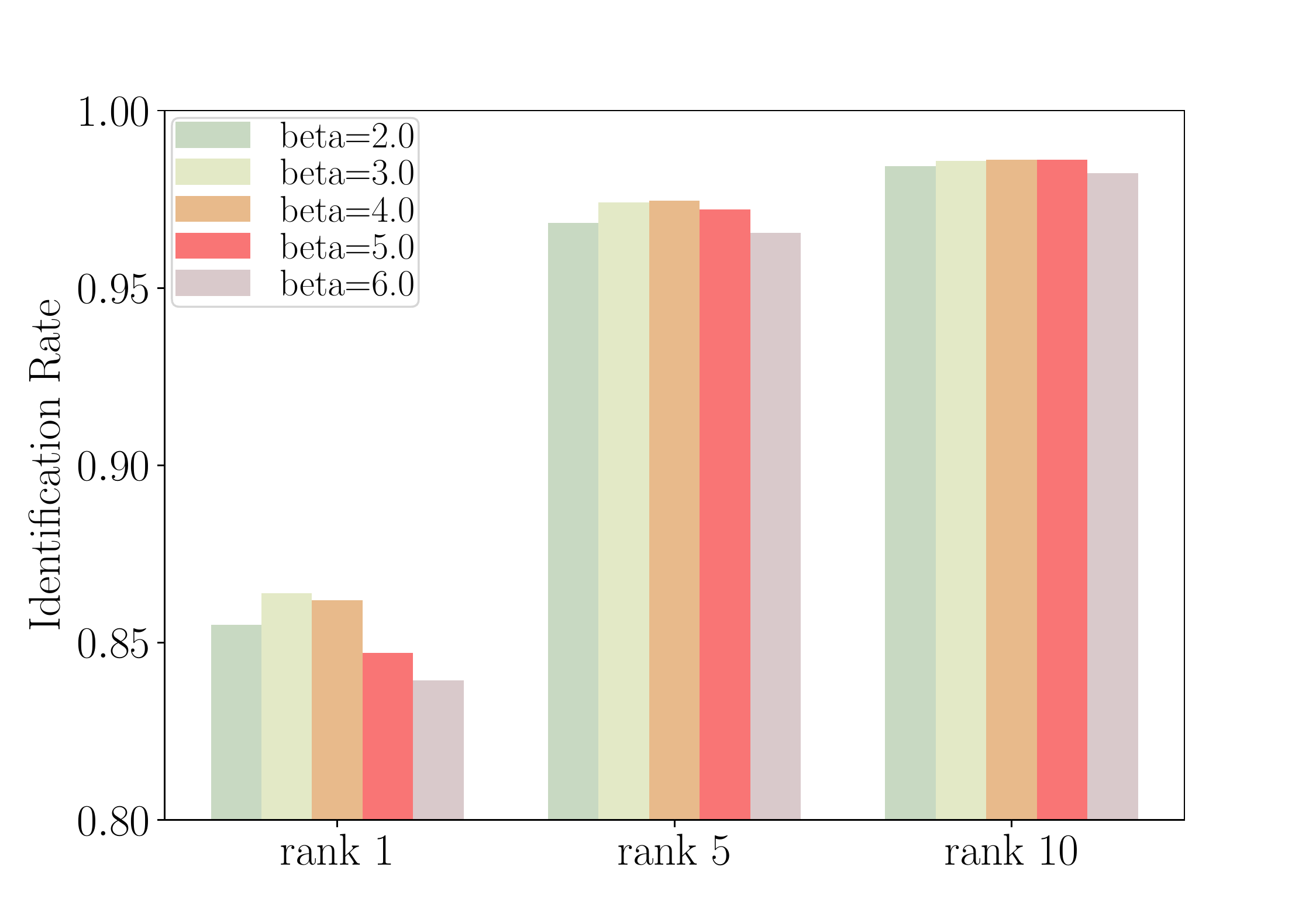}
		\label{fig:beta}
	}
	\subfigure[scale factor in score function $\alpha$]{
		\includegraphics[width=0.31\textwidth]{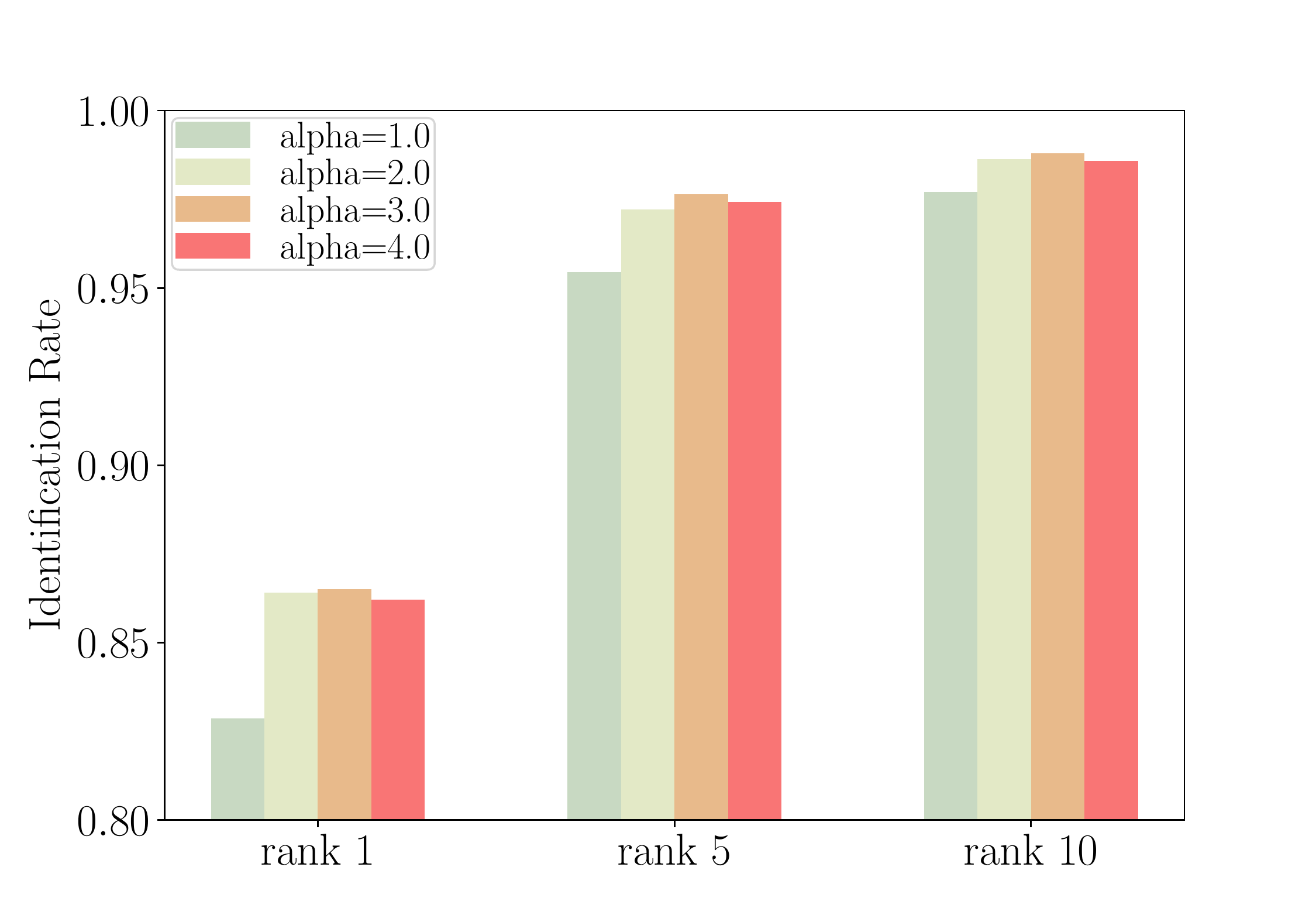}
		\label{fig:alpha}
	}
	\subfigure[transfer loss weight $\lambda$]{
		\includegraphics[width=0.31\textwidth]{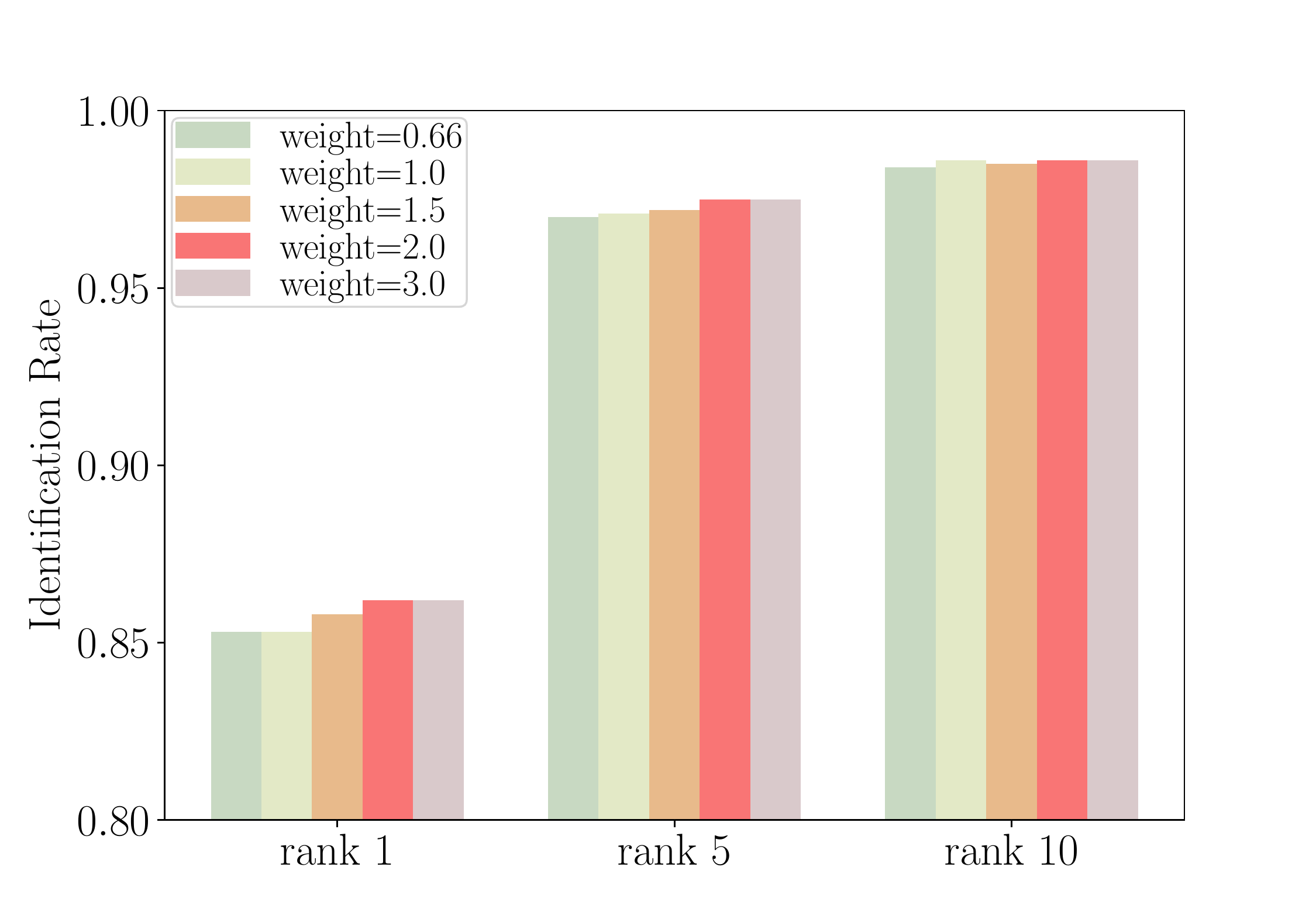}
		\label{fig:weight}
	}
	\caption{The effect of different parameters on the performance of CUHK03 validation set. Here we report Rank-1, 5, 10 results.}
\end{figure*}

In this section, we conduct ablation analysis on the hyper-parameters for our proposed soft DarkRank method, and discuss how they affect the ReID performance.

\paragraph{Contrast $\beta$} Since the rank information only reveals the relative distance between the query and each candidate, it does not provide much details of the absolute distance in the metric space. If the distances of candidates and the query are close, the associated probabilities for the permutations are also close, which makes it hard to distinguish from a good ranking to a bad ranking. So we introduce the contrast parameter $\beta$ to sharpen the differences of the scores. We test different values of $\beta$ on CUHK03 validation set, and find 3.0 is where the model performance peaks. Figure~\ref{fig:beta} shows the details.

\paragraph{Scaling factor $\alpha$} While constraining embeddings on the unit hyper-sphere is the standard setting for metric learning methods in person ReID, a recent work\cite{Ranjan2017L2constrainedSL} shows that small embedding norm may hurt the representation power of embeddings. We compensate this by introducing a scaling factor $\alpha$ and test different values on the CUHK03 validation set. Figure~\ref{fig:alpha} shows the influences on performance of different scaling factors. We choose $\alpha = 3.0$ where the model performance peaks.

\paragraph{Loss weight $\lambda$} During the training process, it is important to balance the transfer loss and the original training loss. We set the loss weight of our transfer loss to 2.0 according to the results in Fig.~\ref{fig:weight}. Note that it also reveals that the performance of our model is quite stable in a large range of $\lambda$.

\subsection{Transfer without Identity}
\begin{table}[th]
	\centering
	\begin{tabular}{ccccc}
		\toprule
		& \multicolumn{2}{c}{Single Query} & \multicolumn{2}{c}{Multiple Query} \\
		\cmidrule{2-3} \cmidrule{4-5}
		Method                    & mAP  & Rank 1 & mAP & Rank 1  \\
		\midrule
		FitNet          & 64.0 & 83.4 & 72.4 & 88.6 \\
		FitNet + DarkRank      & 67.3 & 85.3 & 74.9 & 90.3 \\
		\bottomrule
	\end{tabular}
	\caption{ mAP(\%) and Rank-1 accuracy(\%) on Market1501 of FitNet. We use average pooling features in multi-query test.}
	\label{table:FitNet}
\end{table}

Supervised learning has achieved great success in computer vision, but the majority of collected data  remains unlabeled. In tasks like self-supervised learning\cite{Wang_2015_ICCV}, class level supervision is not available. The supervision signal purely comes from pairwise similarity. Knowledge transfer methods like KD are hard to fit in these cases.  As an  advantage, our method utilize instance level supervision, and thus is available for both supervised and unsupervised tasks. Another well-known instance level method is FitNet\cite{Romero2014FitNetsHF}, which directly matches the embeddings of student and teacher with L2 loss. We compare the transfer performance of FitNet  with and without our DarkRank. As shown in Table.~\ref{table:FitNet}, FitNet achieves similar performance as our method alone. And combined with our method, a significant improvement is achieved. This result further proves that our method utilizes a different kind of information complimenting existing intra-instance methods.


\subsection{Image Retrieval and Clustering Results}
\begin{table}[th]
	\centering
	\begin{tabular}{cccc}
		\toprule
		Method                    & $F_1$     & NMI    & Recall@1   \\
		\midrule
		Student          & 0.153 & 0.461 & 0.311     \\
		DarkRank & 0.168 & 0.483 & 0.340     \\
		\midrule
		Teacher    & 0.172 & 0.484 & 0.367     \\
		\bottomrule
	\end{tabular}
	\caption{ $F_1$, NMI, Recall@1 of DarkRank on CUB-200-2011.}
	\label{table:irc}
\end{table}

The goal of image clustering is to group images into categories according to their visual similarity. And image retrieval is about finding the most similar images in a gallery for a given query image. These tasks rely heavily on the embeddings learnt by model, since the similarity of a image pair is generally calculated based on the Euclidean or Mahalanobis distance between their embeddings. The metrics we adopted for image clustering are $F_1$ and NMI. $F_1$ is the harmonic mean of precision and recall. $F_1 = 2PR/(P+R)$. The Normalized Mutual Information(NMI) reflects the correspondence between candidate clustering $\Omega$ and ground-truth clustering $\mathbb{C}$ of the same dataset. $\text{NMI} = 2I(\Omega, \mathbb{C})/(H(\Omega)+H(\mathbb{C}))$, here $I(\cdot)$ and $H(\cdot)$ are mutual information and entropy, respectively. NMI ranges from 0 to 1, where higher value indicates better correspondence. We choose Recall@1, which is the percentage of returned images belongs to the same category as the query image, as the metric for image retrieval task. The networks and hyper-parameters are as stated in implementation details section. We present the image retrieval and clustering results on CUB-200-2011 in Table.~\ref{table:irc}. The results show our method achieves significant margin in all $F_1$, NMI, Recall@1 metrics. This again shows our method is generally applicable to various kinds of metric learning tasks.

\subsection{Speedup}
\begin{table}[thb]
	\centering
	\begin{tabular}{ccc}
		\toprule
		Model                    & NIN-BN   & Inception-BN \\
		\midrule
		Number of parameters      & 7.6M & 10.3M \\
		Images / Second           & 526  & 178 \\  
		Speedup                   & 2.96 & 1.00 \\
		Rank-1 on CUHK03          & 0.887 & 0.897 \\
		Rank-1 on Market1501    & 0.867 & 0.898 \\
		\bottomrule
	\end{tabular}
	\caption{Complexity and performance comparisons of the student network and teacher network.}
	\label{table:speed}
\end{table}
We summarize the complexity and the performance of the teacher and the student network in Table.~\ref{table:speed}. The speed is tested on Pascal Titan X with MXNet~\cite{chen2015mxnet}. We don't further optimize the implementation for testing. Note that, as the first work that studies knowledge transfer in deep metric learning model, we choose two off-the-shelf network architectures rather than deliberately designing them. Even though, we still achieve a ~3X wall time acceleration with minor performance loss. We believe we can further benefit from the latest network design philosophy~\cite{He2016DeepRL,huang2017densely}, and achieve even better speedup.

\section{Conclusion}
In this paper, we have proposed a new type of knowledge -- cross sample similarities for model compression and acceleration. To fully utilize the knowledge, we have modified the classical listwise rank loss to bridge teacher networks and student networks. Through our knowledge transfer, the student model can significantly improve its performance on various metric learning tasks. Moreover, by combining with other transfer methods which exploit the intra-instance knowledge, the performance gap between teachers and students can be further narrowed. Particularly, without deliberately tuning the network architecture, our method achieves about three times wall clock speedup with minor performance loss with off-the-shelf networks. 
We believe our preliminary work provides a new possibility for knowledge transfer based model acceleration. In the future, we would like to exploit the use of cross sample similarities in more general applications beyond deep metric learning.

\section{Acknowledgments}
This work was supported in part by the National Natural Science Foundation of China (No. 61773375, No. 61375036, No. 61602481, No. 61702510), and in part by the Microsoft Collaborative Research Project.

\bibliography{KD_rank,face-veri,img-retri,person-reid,network,net-prune,quant,distill,metric,l2r}
\bibliographystyle{aaai}
\end{document}